\documentclass[nohyperref]{article}

\usepackage{microtype}
\usepackage{graphicx}
\usepackage{subfigure}
\usepackage{booktabs} 

\usepackage{hyperref}

\usepackage[accepted]{icml2022}

\usepackage{amsmath}
\usepackage{amssymb}
\usepackage{mathtools}
\usepackage{amsthm}

\usepackage[capitalize,noabbrev]{cleveref}

\theoremstyle{plain}

\theoremstyle{definition}

\theoremstyle{remark}

\begin{document}

\twocolumn[
\icmltitle{Paused Agent Replay Refresh}

\icmlsetsymbol{equal}{*}

\begin{icmlauthorlist}
\icmlauthor{Benjamin Parr}{}{me@benjaminparr.com}
\end{icmlauthorlist}



\vskip 0.3in
]




\begin{abstract}

Reinforcement learning algorithms have become more complex since the invention of target networks. Unfortunately, target networks have not kept up with this increased complexity, instead requiring approximate solutions to be computationally feasible. These approximations increase noise in the Q-value targets and in the replay sampling distribution. Paused Agent Replay Refresh (PARR) is a drop-in replacement for target networks that supports more complex learning algorithms without this need for approximation. Using a basic Q-network architecture, and refreshing the novelty values, target values, and replay sampling distribution, PARR gets 2500 points in Montezuma's Revenge after only 30.9 million Atari frames. Finally, interpreting PARR in the context of carbon-based learning offers a new reason for sleep.

\end{abstract}

\section{Introduction}
Reinforcement learning agents interact with an environment to create an experience $(s, a, r, s')$, representing the original state, the action taken, the reward received from the environment, and the new state respectively. An experience replay stores these experiences in the order they were created, allowing the agent to sample past experiences according to some replay sampling distribution \cite{lin1992reinforcement}.

\citet{mnih2015human} trained agents to play Atari using an experience replay with a uniform sampling distribution and a fixed capacity where the oldest experiences are removed to free space to add new experiences. They also introduced a separate Q-network ($Q'$) to generate target Q-values, and whose weights were only updated every $\alpha$ training steps by copying from the current weights of the online Q-network ($Q$). Each training step updates $Q$ by first computing the target Q-value using $Q'$ and the discount factor $\gamma$ according to the rule $target = r + \gamma\ max_{a'} Q'(s', a')$, except when state $s$ is a terminal state in which case the $target = r$. Using this target network instead of the online network to compute the target value increased learning stability by reducing issues caused by the coupled nature of neural network outputs. For example, increasing $Q(s, a)$ tends to also increase $Q(s', a')$. When $Q$ is used to compute the target value instead of $Q'$, this target value tends to also increase, leading to policy divergence or policy oscillation. The target network reduces this issue since $Q'(s', a')$ is unaffected by increasing $Q(s, a)$ until the periodic update of the target network weights.

Target networks have not scaled well with the increasing complexity of reinforcement learning algorithms, causing nosier targets and sampling. For example, adding a recurrent layer to the Q-network means that potentially long sequences of preceding hidden states need to be computed in the training step in order to compute the target value. Approximations avoid this infeasible computation, such as zeroing the hidden state at the start of the sample \cite{hausknecht2015deep}, or using a non-trainable burn-in sequence to allow the Q-network to recover from stale hidden states \cite{kapturowski2018recurrent}. However, input approximations add noise to the computed Q-value targets. Target networks can also add noise to the replay sampling distribution when the distribution is more complex than uniformly random. For example, when the sampling distribution depends on the error between the predicted and target Q-values \cite{schaul2015prioritized}, an experience that originally had a low error might never be sampled, allowing its predicted Q-value to worsen without detection.

\subsection{Sparse Environment Rewards}

Another source of noise from target networks occurs when the target Q-values depend on complex statistics, such as an intrinsic reward for environments with sparse $r$ rewards. For example, in Montezuma's Revenge it takes about 300 frames to collect the key in the first room and get the first points in the game. So, when the environment advances 4 frames per action, the agent does not get a single reward from the environment until completing about 75 actions at minimum, including traversing two ladders, a rope jump, and avoiding an enemy. In order to make progress, the agent must either get unreasonably lucky, or act curiously by pursuing novel states.

\citet{burda2018exploration} defined an intrinsic novelty reward, called Random Network Distillation (RND). RND introduces two new neural networks that take normalized states as inputs. The first neural network is randomly initialized and then fixed. The second neural network is a predictor network that trains to predict the output of the fixed neural network. The novelty reward for a state is its normalized error between the outputs of both networks. The error is normalized so that the novelty reward is on a similar scale regardless of the environment. Finally, the novelty reward of the new state $s'$ is combined with the environment reward $r$ when computing the target value for an experience.

Consider when the agent reaches a new state that is rare and unlike other states it has seen. The novelty reward will generally be larger since the predictor neural network has few, if any, training samples for predicting the output of the fixed neural network. So, the corresponding target value will be larger, causing the agent to want to reach that novel state again. As the state is reached again, the novelty reward decreases since the state is less and less novel. With RND, the agent acts curiously by tending to pursue states that are currently novel.

Both networks take in normalized states as inputs. Normalizing the inputs is important since the fixed neural network can not adjust its weights. Without input normalization, the output of the fixed neural network may have little information about the input, causing the output to be approximately constant across different states. The predictor neural network would then be able to accurately predict the output of the fixed neural network, even for a state that has never been seen before, breaking the goal of the novelty reward calculation.

\section{Paused Agent Replay Refresh}

Paused Agent Replay Refresh (PARR) is a periodic update that handles more complex reinforcement learning algorithms. The replay no longer stores just experiences, but also stores all other online Q-network inputs (e.g. recurrent hidden states), the current target Q-value, and the current sampling distribution parameters. When training, the replay samples according to the sampling distribution parameters, and directly supplies all online Q-network inputs and the target Q-values. So, PARR simplifies the training step since each step updates $Q$ by just using the target Q-value stored in the replay. The computation of Q-value targets is instead moved to the periodic update.

Every $\alpha'$ training steps, PARR pauses all training which temporarily fixes the weights of the online Q-network. Using these temporarily fixed weights, PARR iterates from oldest to newest in the replay, and uses the stored experiences to recompute all other values stored in the replay. After refreshing, all data stored in the replay is as if the current temporarily fixed Q-network was the one that originally generated all the data in the replay. The experiences themselves are not changed when refreshing, and are instead used to refresh all other data without any interaction with the actual environment. After all the data in the replay is refreshed, including the target Q-values and sampling distribution parameters, the agent resumes training.

By setting $\alpha' = \alpha$, PARR becomes a drop-in replacement for target networks in the learning algorithm used by \citet{mnih2015human}. Since their replay sampling distribution is uniform, it is unchanged by replacing the target network with PARR. Also, the target value computed by PARR is identical to the output of the target network that would have been computed during training. This only pertains to experiences already in the replay, and not to new experiences generated after a refresh. Section \ref{Experiment} explores which network to use to compute the target values and replay sample distribution parameters of these new experiences. Using the current online Q-network worked best, and so unlike target networks, PARR does not need a second Q-network.

PARR would increase the wall-clock time of the algorithm used by \citet{mnih2015human}, since it would compute unused Q-value targets. PARR can compensate for this increased periodic computation cost when the learning algorithm is more complex. For example, it removes the need for a non-trainable burn-in period to recover from stale hidden states when the Q-network includes recurrent layers. PARR removes noise in the Q-value targets, and the replay sampling distribution. Finally, PARR enables new algorithms by allowing the computation of complex statistics across the current replay when refreshing without needing noisy approximations.

\subsection{Refreshing Novelty}

\citet{burda2018exploration} used RND with Proximal Policy Optimization (PPO), an on-policy method that does not sample past experiences using a replay. Instead, the most recent experiences are all used to update the policy, and then discarded. Using RND with a replay and target networks causes hard to avoid noise.

First, if the target network only includes the Q-network, then the novelty reward of an experience is unchanged while the experience is in the replay and samplable. The longer the experience stays in the replay, the more stale the novelty reward becomes. An experience that has been in the replay a long time and is soon to be removed from the replay can have a vastly different novelty reward than an exactly identical one that was recently added. So, depending on which experience the agent samples, the agent is told two different target values. Since both experiences are identical, the Q-network can not accurately predict both target values.

This problem can be solved by creating target network versions of both novelty neural networks. So, when using the target networks to compute the target value, the target networks will also recompute the novelty value. However, the novelty statistic is complex and also has normalization parameters. For example, the error between the two novelty neural networks is normalized. \citet{burda2018exploration} maintained a running estimate of the standard deviation of this error. When using a replay, this normalization can be based on just the experiences currently in the replay, thus ignoring experiences that have since been removed from the replay. However, the standard deviation depends on the current novelty network errors. In order to compute the current standard deviation of the network error, the agent must first compute the novelty network error of all experiences in the replay. With PARR, this large standard deviation computation, as well as the computation of the novelty rewards and target values, is done during the refresh.

PARR can simplify refreshing the novelty rewards by using the fact that the novelty rewards are only used for computing target values, which are fixed when training the Q-network. This means that all novelty neural network training and novelty reward computation can be done during the refresh.

Let $R$ be the list of experiences in the replay right before a refresh. During the refresh, the Q-values, novelty reward values, target values and replay sampling distribution of $R$ are all refreshed. To refresh the novelty reward values, first reset the weights of the novelty predictor network, and then train that network from scratch. After training the novelty predictor network, compute the unnormalized novelty network errors for all of $R$, compute the standard deviation of these newly computed novelty network errors, and finally compute the novelty reward by dividing the novelty network errors by the computed standard deviation. These refreshed novelty rewards are then used when refreshing the target values of $R$. After refreshing, the agent resumes simultaneously creating experiences and training the Q-network. Refreshing novelty rewards instead of using target networks ensures the novelty rewards are not stale, and removes noise in computing the normalization parameters.

\section{Experiment}
\label{Experiment}

All experiments ran using Atari 2600 Montezuma's Revenge with sticky actions (sticky action probability of 0.25), even for greedy evaluations \cite{machado2018revisiting}. All experiments use a replay sampling distribution based on the error between the predicted and target Q-values, and completely refresh this replay sampling distribution during each refresh. So, during the refresh, the Q-values, the novelty values, the target values and the replay sampling distribution parameters for all experiences in the replay are refreshed.

The agent was on average able to consistently get 2500 points during evaluation after 30.9 million Atari frames by using PARR. \citet{burda2018exploration} reached 2500 after completing 11\% of their training, which overall used 1.97 billion frames. So PARR used about 7 times fewer frames than the 217 million frames used by \citet{burda2018exploration} to reach 2500 points. PARR was able to achieve this while using the same Q-network architecture as used in \citet{mnih2015human}, with the only change being using a leaky version of the rectifier activation function.

\begin{figure}[ht]
\includegraphics[scale=0.4]{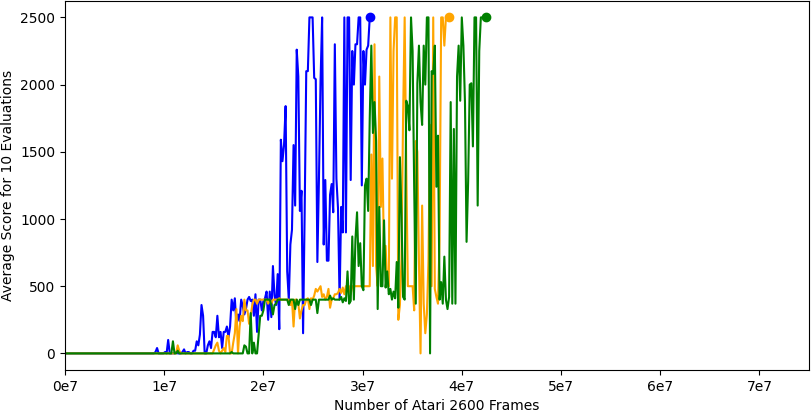}
\vskip 0.25in
\includegraphics[scale=0.4]{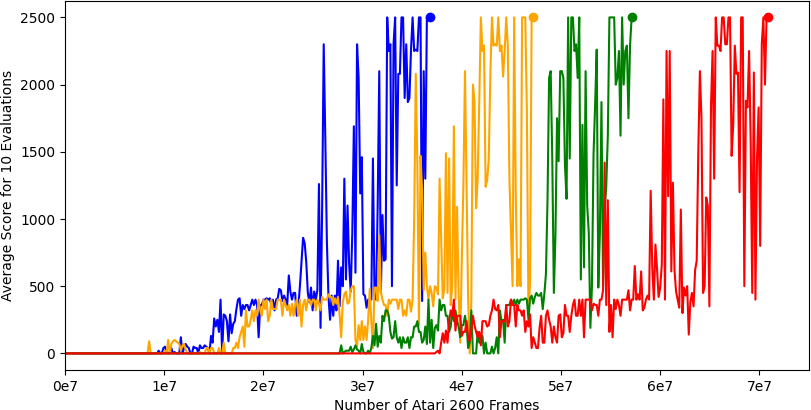}
\caption{Experiments for which network to use to compute the target values and replay sampling distribution parameters for experiences created since the most recent refresh. Three identical trials using the current online Q-network (top), and four identical trials using a copy of the Q-network which was copied after the most recent refresh (bottom).}
\label{results_figure}
\end{figure}

\subsection{Network Architecture}

All experiments were run using the same preprocessing and Q-network architecture as \citet{mnih2015human}. The only network architecture difference is using leaky rectifier activation functions instead of non-leaky rectifier activation functions. Finally, Montezuma's Revenge does not have the flicker issue that some Atari 2600 games have, so each frame is used as is, and does not include taking the element-wise maximum across the current frame and the previous frame.

Specifically, each environment step corresponds to four Atari frames. Every fourth Atari 2600 frame is preprocessed into an 84 x 84 grayscale image. The most recent four preprocessed frames are episodically stacked to form the 84 x 84 x 4 state that is the input to the Q-network. The first preprocessed frame in the episode is repeated to fill in the stack for the first few states of that episode.

The input is then fed into a convolution layer with 32 filters of size 8 x 8 with a stride of 4. That output is fed into a convolution layer with 64 filters of size 4 x 4 with a stride of 2. That output is fed into a convolution layer with 64 filters of size 3 x 3 with a stride of 1. That output is then flattened and fed into a fully-connected layer with 512 outputs. Finally, that output is fed into another fully-connected layer with the same number of actions as the Atari 2600 game (i.e. 18 outputs for Montezuma's Revenge). Each layer is followed by a leaky rectifier activation function.

Finally, just like \citet{mnih2015human}, the environment rewards are clipped between -1 and 1. All non-zero environment rewards in Montezuma's Revenge are $\geq 1$, and so the environment reward is 1.0 when the agent gets points in the game, and 0.0 otherwise.

\subsection{Results}

All experiments were run with refresh frequency $\alpha' = 10,000$ Q-network training steps, identical to the target network update frequency in \citet{mnih2015human}. Also, the agent takes 4 actions between each Q-network training step. Since each action results in 4 Atari 2600 frames, each refresh happens after 160,000 Atari 2600 frames.

The experiments consider two separate ways of computing the target values and replay sampling distribution parameters for experiences created since the most recent refresh. The first experiment using the current online Q-network to compute these new values. The second experiment uses a copy of the Q-network after the most recent refresh to compute these new values. In both experiments, the current online Q-network is used when choosing which action to take next.

After each refresh, the current online Q-network is evaluated by running 10 episodes using a greedy policy. The average of all 10 evaluation episodes is shown in Figure \ref{results_figure}. Each evaluation is non-deterministic because of sticky actions. Each trial is stopped after the tenth time all evaluation episodes resulted in 2500 points.

The first refresh happens immediately after the burn-in phase and before any training. The burn-in phase creates 25,000 experiences (100,000 Atari 2600 frames). After that, every refresh happens after an additional 40,000 experiences (160,000 Atari frames). On average, using the current online Q-network first consistently got 2500 points during evaluation after 30.9 million Atari frames. On average using the copy of the Q-network did so after 47.8 million frames. The latter experiment was run for an additional trial because of the larger variance in its results.

This choice of network only affects the experiences created after refresh. The replay capacity is $2^{20}$ = 1,048,576 experiences. Immediately before a refresh, the experiment affects 40,000 experiences, which is about 3.8\% of a full replay. Immediately after a refresh, there is no difference between the two experiments. Despite this low percent of affected experiences, using the current online Q-network appears to perform better. Using the current online Q-network is also simpler algorithmically. When using the copied Q-network approach, a newly created experience has to be fed into both the current online Q-network as well as the copied Q-network. Using just the current online Q-network means only one Q-network is needed.

\begin{table}[t]
\begin{center}
\caption{Results for the three identical trials using the current online Q-network. Shows when each trial achieved 2500 points in all 10 evaluations for the first time, as well as the tenth time. All entries are formatted as: Number of Atari Frames (Number of Refreshes).}
\begin{small}
\begin{sc}
\begin{tabular}{cc}
\toprule
$1^{st}$ All 2500 Evaluation & $10^{th}$ All 2500 Evaluation \\
\midrule
24.74 million\ \ \ (155) & 30.82 million\ \ \ (193) \\
32.90 million\ \ \ (206) & 38.82 million\ \ \ (243) \\
34.98 million\ \ \ (219) & 42.50 million\ \ \ (266) \\
\bottomrule
\end{tabular}
\end{sc}
\end{small}
\vskip 0.35in

\caption{Results for the four identical trials using a copy of the Q-network which was copied after the most recent refresh. Shows when each trial achieved 2500 points in all 10 evaluations for the first time, as well as the tenth time. All entries are formatted as: Number of Atari Frames (Number of Refreshes).}
\vskip 0.2in
\begin{small}
\begin{sc}
\begin{tabular}{cc}
\toprule
$1^{st}$ All 2500 Evaluation & $10^{th}$ All 2500 Evaluation \\
\midrule
32.58 million\ \ \ (204) & 36.90 million\ \ \ (231) \\
42.02 million\ \ \ (263) & 47.30 million\ \ \ (296) \\
50.82 million\ \ \ (318) & 57.22 million\ \ \ (358) \\
65.70 million\ \ \ (411) & 70.98 million\ \ \ (444) \\
\bottomrule
\end{tabular}
\end{sc}
\end{small}
\end{center}
\vskip -0.15in
\end{table}

\subsection{Hardware}

Each experiment was run on a consumer desktop machine with a single GPU, and used a replay capacity of $2^{20}$ = 1,048,576. So, each replay contained $4.8\%$ more experiences than the replay in \citet{mnih2015human} which used a capacity of 1,000,000. The replay in a PARR experiment also stored more data than \citet{mnih2015human}, and so each PARR experiment required more RAM. Specifically, each experiment used 17.2G of RAM in order to run.

\section{Discussion}

PARR is described here as a serial process for clarity. The refresh can be optimized and parts done in parallel without compromising the end goal of having the replay contain data as if all experiences were generated by the current online Q-network. Refreshing without pausing training may also be possible. However, PARR is periodic like target network updates for the same reason, namely to avoid policy divergence and policy oscillations caused by the coupling of neural network outputs.

PARR for novelty is described here as training the novelty network from scratch during refresh. This ensures the novelty network is only trained on experiences actually in the replay at the time of refresh, and ignores any experiences that were removed right before the refresh. By relaxing this reasoning, parts of the novelty reward training can be done alongside the Q-network training in order to speed up the refresh.

\subsection{Future Work}

All experiments used the same basic Q-network architecture as \citet{mnih2015human}. Since then, there have been advances in the architecture itself that could be added to the learning algorithm. For example, a recurrent layer could give the agent short-term memory. Refreshing the hidden states of such a layer is straightforward given the approach described in this paper, since the hidden states can be refreshed sequentially alongside the Q-values.

All experiments converged to getting 2500 points in Montezuma's Revenge. After getting the first key, the agent has the option of going left or right. The room to the left is a room with deadly periodic laser gates and so is more difficult than going right. So, even when the agent discovered going left first, it still converged to going right, since that path is easier to find novel states and environment rewards. After going right, the agent gets the sword in the room below, then backtracks to use the sword on an enemy in the previous room. At that point, the only way to get more points is to get past deadly periodic laser gates in the room to the right of the room with the sword. Even though the agent occasionally gets past the first few laser gates, it does not learn to do so consistently. This is because, even though getting past the first few laser gates is very novel in theory, the novelty reward is relatively low. Instead, while in that room, the novelty reward is more dictated by whether or not the laser gates are currently visible instead of the location of the player in the room. This is a problem with the novelty calculation that requires future work to fix.

\subsection{Creative Interpretation}

Paused Agent Replay Refresh offers a new learning-based justification for sleep. When we sleep, we also pause and stop interacting with the world. PARR allows the agent to understand its experience from the perspective of who it is now, instead of who it was when the agent created the experience. Perhaps sleep does similar for us. This perspective shift requires computation, and therefore time, and so it can't be instant like copying weights to a target network. With the target network approach, agents must continuously train with all but a wink of rest. Perhaps learning machines could use some sleep too.

\vskip 0.3in
\bibliography{paper}

\begin{thebibliography}{7}
\providecommand{\natexlab}[1]{#1}
\providecommand{\url}[1]{\texttt{#1}}
\expandafter\ifx\csname urlstyle\endcsname\relax
  \providecommand{\doi}[1]{doi: #1}\else
  \providecommand{\doi}{doi: \begingroup \urlstyle{rm}\Url}\fi

\bibitem[Burda et~al.(2018)Burda, Edwards, Storkey, and
  Klimov]{burda2018exploration}
Burda, Y., Edwards, H., Storkey, A., and Klimov, O.
\newblock Exploration by random network distillation.
\newblock \emph{arXiv preprint arXiv:1810.12894}, 2018.

\bibitem[Hausknecht \& Stone(2015)Hausknecht and Stone]{hausknecht2015deep}
Hausknecht, M. and Stone, P.
\newblock Deep recurrent q-learning for partially observable mdps.
\newblock In \emph{2015 aaai fall symposium series}, 2015.

\bibitem[Kapturowski et~al.(2018)Kapturowski, Ostrovski, Quan, Munos, and
  Dabney]{kapturowski2018recurrent}
Kapturowski, S., Ostrovski, G., Quan, J., Munos, R., and Dabney, W.
\newblock Recurrent experience replay in distributed reinforcement learning.
\newblock In \emph{International conference on learning representations}, 2018.

\bibitem[Lin(1992)]{lin1992reinforcement}
Lin, L.-J.
\newblock \emph{Reinforcement learning for robots using neural networks}.
\newblock Carnegie Mellon University, 1992.

\bibitem[Machado et~al.(2018)Machado, Bellemare, Talvitie, Veness, Hausknecht,
  and Bowling]{machado2018revisiting}
Machado, M.~C., Bellemare, M.~G., Talvitie, E., Veness, J., Hausknecht, M., and
  Bowling, M.
\newblock Revisiting the arcade learning environment: Evaluation protocols and
  open problems for general agents.
\newblock \emph{Journal of Artificial Intelligence Research}, 61:\penalty0
  523--562, 2018.

\bibitem[Mnih et~al.(2015)Mnih, Kavukcuoglu, Silver, Rusu, Veness, Bellemare,
  Graves, Riedmiller, Fidjeland, Ostrovski, et~al.]{mnih2015human}
Mnih, V., Kavukcuoglu, K., Silver, D., Rusu, A.~A., Veness, J., Bellemare,
  M.~G., Graves, A., Riedmiller, M., Fidjeland, A.~K., Ostrovski, G., et~al.
\newblock Human-level control through deep reinforcement learning.
\newblock \emph{nature}, 518\penalty0 (7540):\penalty0 529--533, 2015.

\bibitem[Schaul et~al.(2015)Schaul, Quan, Antonoglou, and
  Silver]{schaul2015prioritized}
Schaul, T., Quan, J., Antonoglou, I., and Silver, D.
\newblock Prioritized experience replay.
\newblock \emph{arXiv preprint arXiv:1511.05952}, 2015.

\end{thebibliography}
\bibliographystyle{icml2022}

\end{document}